\documentclass[11pt]{article}
\usepackage{acl2015}
\usepackage{times}
\usepackage{url}
\usepackage{latexsym}
\usepackage{graphicx} 
\usepackage{booktabs}
\usepackage{hyperref}
\usepackage{natbib}
\title{Traversing Emotional Landscapes and Linguistic Patterns in Bernard-Marie Koltès' Plays: An NLP Perspective}

\author{Arezou Zahiri Pourzarandi \\
  {\tt a.zahiri@student.art.ac.ir} \\\And
  Farshad Jafari \\
  {\tt farshad.jafari@gatech.edu} \\}

\date{}

\begin{document}

\maketitle

\begin{abstract}
This study employs Natural Language Processing (NLP) to analyze the intricate linguistic and emotional dimensions within the plays of Bernard-Marie Koltès, a central figure in contemporary French theatre. By integrating advanced computational techniques, we dissect Koltès' narrative style, revealing the subtle interplay between language and emotion across his dramatic oeuvre. Our findings highlight how Koltès crafts his narratives, enriching our understanding of his thematic explorations and contributing to the broader field of digital humanities in literary analysis.
\end{abstract}
\section{Introduction}

Bernard-Marie Koltès (1948–1989) stands as a pivotal figure in late 20th-century French theatre, renowned for his profound explorations of themes such as isolation, power dynamics, communication, and existential despair \citep{conte2017theatre}. His works are characterized by a rich psychological depth, complex characters, and the exploration of marginalized or liminal spaces that often address postcolonial identity and questions of national belonging \citep{kleppinger2007questioning}. Koltès' ability to blend poetic language with stark realism allows his plays to transcend simple storytelling, making them a vehicle for both philosophical inquiry and social commentary.

Koltès' dramatic compositions, such as Dans la Solitude des Champs de Coton, La Nuit Juste Avant les Forêts, and Combat de Nègre et de Chiens, focus intensely on the interaction between characters from different backgrounds, revealing the tensions and desires that shape human relationships. His use of language serves as a crucial tool in this regard, shaping the atmosphere, pacing, and emotional texture of each play. Despite Koltès' profound influence on modern French theatre, the complexity of his language and the emotional depth of his characters have made comprehensive literary analysis challenging using traditional methods.

\subsection{Research Motivation}
The challenge of interpreting Koltès’ works lies in their intricate linguistic structures and the emotionally charged nature of the dialogue. The interplay of themes like desire, identity, and violence, along with Koltès' stylistic use of language, invites a more nuanced analytical approach that extends beyond conventional literary critique. Despite the rich body of scholarship on his work, there remains a significant gap in understanding how language functions at a deeper, more structural level within his plays—particularly in how it drives emotional engagement and dramatic tension.

To address this gap, we employ state-of-the-art Natural Language Processing (NLP) techniques to analyze Koltès’ works from a computational perspective. By leveraging tools such as sentiment analysis, emotion detection, and linguistic feature extraction, we aim to uncover the patterns and structures that underpin the psychological and emotional depth of Koltès’ plays. This approach allows for a quantitative and systematic exploration of Koltès' use of language, providing novel insights into the emotional trajectories of his characters and the dramatic tension that defines his narratives.

\subsection{Objectives}
Our research focuses on three core objectives, aiming to bridge the gap between computational linguistics and traditional literary analysis:

\textbf{Linguistic Patterns}: We seek to explore the distinctive linguistic features of Koltès' texts, such as word frequency, vocabulary richness, and the Type-Token Ratio (TTR). These analyses will shed light on Koltès' stylistic choices, revealing how his use of language contributes to the thematic and atmospheric elements of his plays.

\textbf{Emotional Trajectories}: By applying sentiment analysis and emotion detection models, we will map the evolving emotional landscapes within Koltès' narratives. This approach will allow us to trace the emotional peaks and valleys that contribute to the overall dramatic tension, offering a deeper understanding of the characters' psychological depth and the emotional impact on the audience.

\textbf{Dramatic Tension}: Through computational analysis of sentiment and emotional expressions over the course of the plays, we aim to quantify the dramatic tension and pacing in Koltès' work. This analysis will help to elucidate the mechanisms through which Koltès builds suspense and engages the audience, further enhancing our understanding of his dramatic artistry.

\subsection{Scope of the Study}
This study focuses on Koltès' most renowned works, including Dans la Solitude des Champs de Coton, La Nuit Juste Avant les Forêts, and Combat de Nègre et de Chiens.

\textbf{{Dans la Solitude des Champs de Coton (In the Solitude of the Cotton Fields)}}: This play presents a tense, nocturnal encounter between a Dealer and a Client, set against the backdrop of an unspecified, desolate landscape reminiscent of cotton fields. The dialogue-driven narrative explores themes of desire, commerce, and the inherent violence in transactions, whether they be emotional, physical, or commercial. The play's ambiguous setting and poetic language serve to universalize the confrontation, making it a metaphor for broader human experiences and interactions.

\textbf{{La Nuit Juste Avant les Forêts (The Night Just Before the Forests)}}: A monologue that delves into the psyche of an unnamed immigrant man wandering the streets, this play is a poignant exploration of loneliness, alienation, and the desperate human need for connection. The protagonist's fragmented speech reflects his disjointed thoughts and the discontinuities in his life, creating a powerful narrative that challenges the audience to confront the realities of exclusion and the search for identity in an indifferent world.

\textbf{{Combat de Nègre et de Chiens (Black Battles with Dogs)}}: Set in a remote construction site in an unnamed African country, this play examines the complex dynamics of colonialism, racism, and the clash of cultures through the interactions of four characters: two French engineers, an African watchman, and a woman who arrives from France. The play's intense dialogues and dramatic confrontations reveal the underlying prejudices, desires, and fears of the characters, offering a critical look at the lingering impacts of colonialism and the nature of human conflict.

In sum, our work presents a novel approach to literary analysis, merging advanced NLP techniques with traditional dramaturgical study. By focusing on linguistic patterns, emotional trajectories, and dramatic tension, we seek to deepen the understanding of Bernard-Marie Koltès' work, providing a framework for analyzing other dramatic texts in a similarly nuanced and computationally informed manner.

\section{Background}
In recent years, the study of emotion dynamics and sentiment analysis in literary texts has become increasingly prominent, leveraging computational techniques to explore the emotional trajectories of characters and narratives. Sentiment analysis tools have advanced significantly, allowing for the examination of emotional arcs across entire novels or dramatic works. For example, emotion dynamics in literary novels have been studied by distinguishing the emotional arcs of characters from the overall narrative, revealing key insights into how characters’ emotions evolve independently from the broader story \citep{vishnubhotla2024emotion}. However, the scope of these studies has often been limited to relatively straightforward lexicon-based methods, which can fall short in capturing the intricate emotional relationships and interactions between characters.

A substantial body of work has focused on applying sentiment analysis to dramatic texts, where character emotions play a pivotal role in defining the structure and tension of the narrative. In some works, word-emotion lexicons have been employed to map characters’ emotional landscapes, particularly in tragedies where emotional contrast drives the plot \citep{yavuz2020analyses}. Such methods have helped position protagonists and antagonists in emotional space but are constrained by the static nature of lexicon-based approaches that do not fully capture temporal shifts in emotion or the complexity of interactions between multiple characters.

Another approach has explored character-to-character sentiment analysis, especially in Shakespearean plays, where changes in sentiment between characters reflect the unfolding of the drama \citep{nalisnick2013character}. This method highlights the dynamic aspect of character interactions, but traditional sentiment analysis tools often fail to fully capture the nuanced emotional exchanges due to the limitations of older models. In a similar vein, the extraction of social networks from literature, which models relationships based on dialogues and interactions, has been successfully applied to fictional texts \citep{elson2010extracting}. However, these methods have primarily relied on rule-based or shallow learning techniques, limiting their capacity to model more abstract and implicit emotional undercurrents that are crucial in complex narratives like those of Koltès.

Some studies have extended these analyses by constructing sentiment networks that track emotional exchanges between characters across an entire play or novel. For instance, sentiment networks extracted from Shakespeare’s plays have shown how emotional polarities between characters develop and shift over time, adding a layer of depth to the understanding of character relationships \citep{nalisnick2013extracting}. Similarly, character networks in dramatic works have been mapped using graph-based approaches, revealing structural antagonisms that align with traditional literary interpretations \citep{yavuz2020analyses}. These methods, while insightful, are often constrained by their reliance on predefined emotional categories, which limits their ability to capture the full range of emotional variation present in complex, modern works.

There have also been efforts to integrate deeper sentiment analysis into historical texts, such as German plays, where sentiment analysis techniques were adapted to analyze character emotions and relationships \citep{schmidt2018toward}. However, such studies have typically employed older lexicon-based techniques or linear models, which are not well-suited for the nuanced, multi-dimensional emotion dynamics seen in more modern and diverse literary works like those of Koltès. Moreover, mapping emotions in stories using tools like EMOFIEL has provided useful frameworks for visualizing character relationships and their emotional exchanges, though these tools have been primarily lexicon-based and limited in their ability to adapt to non-traditional narratives \citep{jhavar2018emofiel}.

While lexicon-based methods have been highly useful in earlier research, recent advancements in deep learning and natural language processing have opened new avenues for exploring the complexities of emotion and sentiment in literature. For example, more recent work on emotion arcs in text streams has demonstrated the effectiveness of using neural models to capture high-quality emotion arcs that reflect the nuanced emotional shifts within a narrative \citep{teodorescu2022frustratingly}. These approaches outperform traditional lexicon-based models by generating more accurate representations of characters’ emotional trajectories over time.

Despite these advancements, existing research has largely focused on well-known works such as Shakespeare or genre-specific texts like German historical plays, leaving a significant gap in the application of these models to more contemporary or avant-garde works, such as those of Bernard-Marie Koltès. Koltès’ works are characterized by their rich, emotionally charged dialogues and non-linear narrative structures, which present a unique challenge for traditional sentiment analysis models. The complexity and subtlety of emotional expression in his works necessitate more sophisticated models that can go beyond basic lexicon-based sentiment mapping.

In our research, we address these gaps by applying cutting-edge deep learning models to analyze the emotional dynamics in Koltès' plays. Unlike previous approaches that rely heavily on static lexicons or shallow learning techniques, we employ advanced neural networks that can model the intricate and evolving emotional states of characters in real-time. These models not only capture the explicit emotional expressions in the dialogue but also infer deeper, implicit sentiments that are crucial to understanding the layered emotional landscape of Koltès' narratives.

\section{Methodology}
To prepare the texts of Bernard-Marie Koltès' plays for analysis, we undertook a systematic approach to ensure the integrity and quality of the data for NLP processing. The methodology encompassed text extraction, cleaning, segmentation, and the application of various NLP tools and techniques for comprehensive linguistic and emotional analysis.
The codes used for this analysis are all available at the \href{https://github.com/frshdjfry/NLP-in-Dramatic-Literature}{GitHub repository}.

\subsection{Text Extraction and Preprocessing}
Text Extraction: We utilized the French version of Tesseract OCR to digitize printed pages of Koltès' plays, converting them into machine-readable text.
Cleaning: The extracted texts underwent cleaning to remove OCR errors, irrelevant formatting, and non-textual elements to ensure the purity of the data for analysis.
Text Segmentation
Given the theatrical nature of the texts, we segmented the plays into 150-word units, approximating one minute of stage time, to analyze linguistic and emotional content in manageable, performance-relevant chunks.

\subsection{Linguistic Analysis}
Using the spaCy library configured for French, we performed a linguistic analysis that included:

\textbf{Word Frequency and Word Clouds}: Identification of the most common words and visualization using WordCloud in matplotlib to highlight key thematic elements.
\textbf{Vocabulary Richness}: Computation of lexical diversity to assess the linguistic complexity and stylistic variance across the plays.

\subsection{Sentiment Analysis}
We employed a sentiment analysis pipeline built with Hugging Face's transformers library, specifically using the tblard/tf-allocine model fine-tuned for French texts \citep{Blard2020}. This allowed us to assess the sentiment of each text segment, providing insights into the plays' emotional arcs and dramatic tension.

\subsection{Emotion Detection}
To capture a nuanced spectrum of emotions, we used the bhadresh-savani/bert-base-uncased-emotion model via Hugging Face's pipeline for text classification \citep{tunstall2022natural}. This model returned scores for multiple emotions, enabling a detailed emotional profile of each segment.

\subsection{Aggregation and Analysis}
For both sentiment and emotion analyses, we averaged the scores within each 150-word segment to obtain a representative emotional or sentiment value for that "minute" of the play. This approach allowed us to plot the progression of emotional intensity and sentiment valence throughout the plays, offering a novel perspective on Koltès' dramaturgy and thematic layering.

This methodological framework combines traditional literary analysis with cutting-edge NLP techniques, offering a multifaceted exploration of Koltès' works that bridges the gap between qualitative literary insights and quantitative linguistic and emotional patterns.
\section{Analysis and Findings}

\subsection{Dans la Solitude des Champs de Coton}

The analysis of "Dans la Solitude des Champs de Coton" reveals significant insights into the linguistic and emotional landscape of the play. The Type-Token Ratio (TTR) of 0.4919 indicates a diverse vocabulary, reflecting the complex interplay of themes such as desire, identity, and the human-animal dichotomy.

\begin{table}[!htb]
    \centering
    \begin{tabular}{ll}
        \toprule
        Word & Count \\
        \midrule
        homme (human) & 52 \\
        désir (desire) & 39 \\
        heure (hour) & 30 \\
        animal & 26 \\
        temps (time) & 18 \\
        point (place) & 18 \\
        vous (you) & 16 \\
        regard (look) & 15 \\
        main (hand) & 14 \\
        froid (cold) & 14 \\
        \bottomrule
    \end{tabular}
    \caption{Vocabulary Analysis in "Dans la Solitude des Champs de Coton"}
    \label{table:vocab-analysis-dans}
\end{table}

\begin{figure*}[!htb]
    \centering
    \includegraphics[width=\textwidth]{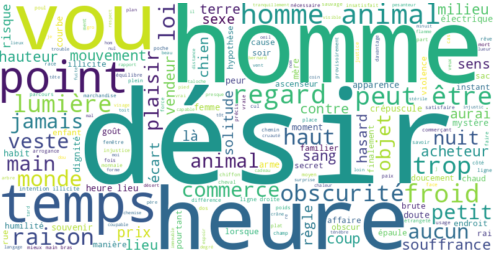}
    \caption{Wordcloud of Most Frequent Terms in "Dans la Solitude des Champs de Coton"}
    \label{fig:wordcloud-dans}
\end{figure*}

\textbf{Desire}: The frequent use of "désir" (desire) without specifying its object underscores the play's exploration of unarticulated desires, acting as a hidden driving force behind the characters' interactions. This theme is visually represented in the word cloud in Figure \ref{fig:wordcloud-dans}.

\textbf{Human and Animal}: The juxtaposition of "homme" (human) and "animal" within the text highlights the blurred lines between human and animalistic instincts, suggesting a fundamental equality and shared primal nature, as seen in the word cloud (Figure \ref{fig:wordcloud-dans}).

\textbf{Time and Place}: The recurring references to "heure" (hour), "temps" (time), and "point" (place) emphasize the play's abstract setting, devoid of specific temporal or spatial anchors, thus universalizing its themes.

\textbf{Coldness}: The motif of "froid" (cold) permeates the play, adding to its chilling atmosphere and possibly symbolizing the emotional coldness and existential isolation of the characters.

The dramatic tension intensifies in the final third of the play, where dialogue becomes increasingly terse, reflecting a climax in conflict. This escalation is quantitatively captured through sentiment analysis over time, as shown in Figure \ref{fig:sentiment-time-dans}.

The emotional analysis, depicted in Figures \ref{fig:emotion-time-dans} and \ref{fig:emotion-percentage-dans}, further delineates the emotional fluctuations throughout the play, with a notable increase in "sadness" and "fear" towards the end, signifying a crescendo in dramatic tension and existential dread.

\subsection{La Nuit Juste Avant les Forêts}

\begin{figure*}[!htb]
    \centering
    \includegraphics[width=\textwidth]{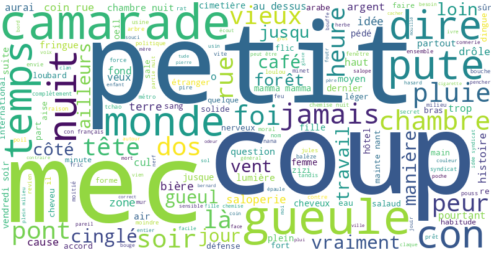}
    \caption{Wordcloud of Most Frequent Terms in "La Nuit Juste Avant les Forêts"}
    \label{fig:wordcloud-nuit}
\end{figure*}

\textbf{Vocabulary Richness}: The Type-Token Ratio of 0.3624, as shown in Table \ref{table:vocab-analysis-nuit}, indicates a more limited vocabulary range, possibly due to the monologue format. The word cloud in Figure \ref{fig:wordcloud-nuit} visually represents the most frequent terms, with words like "nuit" (night) and "chambre" (room) anchoring the narrative in a nocturnal, urban setting, reflecting the protagonist's sense of isolation.

\begin{table}[!htb]
    \centering
    \begin{tabular}{ll}
        \toprule
        Word & Count \\
        \midrule
        nuit (night) & 40 \\
        mec (guy) & 28 \\
        con (idiot) & 27 \\
        chambre (room) & 27 \\
        coup (shot) & 27 \\
        camarade (comrade) & 26 \\
        rue (street) & 25 \\
        gueule (mouth) & 25 \\
        monde (world) & 25 \\
        pute (whore) & 25 \\
        \bottomrule
    \end{tabular}
    \caption{Vocabulary Analysis in "La Nuit Juste Avant les Forêts"}
    \label{table:vocab-analysis-nuit}
\end{table}

\textbf{Narrative Focus}: The contrast between "chambre" and "rue" (street), prominent in the vocabulary analysis (Table \ref{table:vocab-analysis-nuit}), underscores the protagonist's longing for shelter against his reality of wandering the streets. The narrative oscillates between past and present, with the protagonist recounting various encounters marked by themes of rejection and camaraderie, which is further elucidated through the emotional lexicon depicted in the word cloud (Figure \ref{fig:wordcloud-nuit}).

\textbf{Emotional Landscape}: The sentiment analysis over time, illustrated in Figure \ref{fig:sentiment-time-nuit}, suggests a narrative arc characterized by fluctuating emotions, with significant moments of "sadness" and "anger" intensifying towards the end. This reflects the protagonist's increasing desperation and anger, highlighting the emotional depth and complexity of the monologue. The emotion analysis over time (Figure \ref{fig:emotion-time-nuit}) and the percentage distribution of emotions (Figure \ref{fig:emotion-percentage-nuit}) provide a quantitative insight into the varying emotional intensities throughout the play, showcasing the nuanced emotional landscape that underpins the narrative.

\subsection{Combat de Nègre et de Chiens}

\begin{figure*}[!htb]
    \centering
    \includegraphics[width=\textwidth]{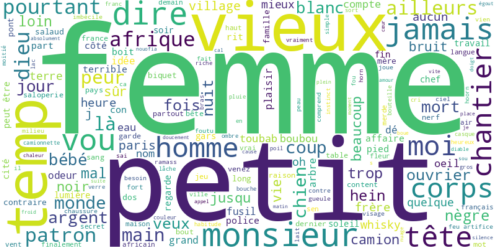}
    \caption{Wordcloud of Most Frequent Terms in "Combat de Nègre et de Chiens"}
    \label{fig:wordcloud-combat}
\end{figure*}

\textbf{Linguistic Patterns}: The Type-Token Ratio (TTR) of 0.3561, detailed in Table \ref{table:vocab-analysis-combat}, indicates focused thematic exploration within the play, particularly through the recurrent use of "femme" (woman), highlighting the themes of objectification and mystification in a male-dominated environment. The word cloud in Figure \ref{fig:wordcloud-combat} visually encapsulates these and other thematic terms, providing an immediate impression of the play's linguistic landscape.

\begin{table}[!htb]
    \centering
    \begin{tabular}{ll}
        \toprule
        Word & Count \\
        \midrule
        femme (woman)& 74 \\
        vieux (old)& 67 \\
        temps (time) & 51 \\
        monsieur (sir)& 49 \\
        dire (to say) & 46 \\
        jamais (never) & 43 \\
        chantier (construction site) & 38 \\
        corps (body) & 37 \\
        tête (head) & 37 \\
        afrique (africa)& 34 \\
        \bottomrule
    \end{tabular}
    \caption{Vocabulary Analysis in "Combat de Nègre et de Chiens"}
    \label{table:vocab-analysis-combat}
\end{table}

\textbf{Setting and Conflict}: The setting is vividly established through the term "chantier" (construction site), signifying the backdrop for the unfolding drama and cultural clashes, as seen in both the vocabulary analysis (Table \ref{table:vocab-analysis-combat}) and the word cloud (Figure \ref{fig:wordcloud-combat}). The frequent mentions of "corps" (body) and "tête" (head) in the text not only underscore the play's physical conflicts but also allude to deeper psychological battles.

\textbf{Character Dynamics}: The formal address of "monsieur" (sir) and the significant focus on "femme" shed light on the gender dynamics and power structures central to the narrative. The word cloud (Figure \ref{fig:wordcloud-combat}) further illustrates these dynamics, emphasizing the pivotal role of the female presence in escalating the tension among the male characters.

\textbf{Emotional and Sentiment Analysis}: The sentiment analysis over time, depicted in Figure \ref{fig:sentiment-time-combat}, illustrates the fluctuating emotional states throughout the play, capturing moments of tension, conflict, and resolution. Complementing this, the emotion analysis over time (Figure \ref{fig:emotion-time-combat}) and the percentage distribution of emotions (Figure \ref{fig:emotion-percentage-combat}) quantitatively dissect the emotional fabric of the narrative, revealing the complexities of the characters' interactions and the undercurrents of their psychological landscapes.

\subsection{Comparative Analysis of Koltès' Plays}
Based on our findings using NLP techniques, we conducted a comparative analysis of Koltès' plays, focusing on vocabulary richness, recurring themes, and emotional dynamics. Through the analysis of metrics such as Type-Token Ratios (TTR), word frequency, and sentiment trajectories, we observed distinct linguistic and emotional patterns across his works. These insights highlight how Koltès’ unique use of language and emotion contributes to the dramatic tension in each play.

\subsubsection{Vocabulary Richness and Linguistic Diversity}
Dans la Solitude des Champs de Coton exhibits a Type-Token Ratio (TTR) of 0.4919, indicating a relatively high vocabulary diversity, which may suggest complex character interactions and a wide range of explorations. In contrast, La Nuit Juste Avant les Forêts and Combat de Nègre et de Chiens have lower TTRs of 0.3624 and 0.3561, respectively, possibly due to the focused narrative scope or the monologue format in La Nuit, which might limit linguistic variety.

\subsubsection{Recurring Themes Through Word Frequency}
The word "désir" (desire) appears 39 times in Dans la Solitude, underlining the play's central theme of unspoken desires and transactions. This theme is less pronounced in the other two plays, where the focus shifts to more immediate physical and emotional experiences, as seen with the frequent mention of "nuit" (night) 40 times in La Nuit and "femme" (woman) 74 times in Combat, highlighting the plays' settings and central motifs.
Emotional Trajectories and Dramatic Tension
Sentiment analysis over time segments reveals a notable increase in dramatic tension towards the end of Dans la Solitude, with shorter, more intense exchanges between characters. This pattern contrasts with La Nuit, where emotional intensity is more evenly distributed, punctuated by peaks of "sadness" and "anger" towards the narrative's climax. Combat shows a consistent thematic focus on conflict and power dynamics, reflected in the sustained presence of "anger" and "fear" throughout the play.

\subsubsection{Emotional Composition}
The emotion analysis presents a varied emotional landscape across the plays. For example, Dans la Solitude shows a balanced mix of "anger" (42.8\%) and "fear" (22.9\%), suggesting a narrative fraught with conflict and uncertainty. La Nuit, however, has moments of "sadness" peaking towards its conclusion, indicating a crescendo of personal despair. In contrast, Combat is dominated by "anger", reflecting its themes of confrontation and cultural clash.

\subsubsection{Synthesis}
These comparative insights reveal Koltès' ability to navigate a broad spectrum of human emotions and experiences through his diverse linguistic choices. While Dans la Solitude delves into the abstract complexities of desire and human interaction, La Nuit offers a more intimate look at personal anguish and societal alienation. Combat, on the other hand, confronts the raw realities of power, violence, and cultural tensions. The varying TTRs and the dominant emotional tones across the plays underscore Koltès' versatile narrative style, from the introspective and philosophical to the starkly realistic and confrontational.

\section{Discussion}
\subsection{Literary Implications}
The NLP-driven analysis of Bernard-Marie Koltès' plays offers profound insights into his dramatic oeuvre and its place within modern theatre. By quantitatively dissecting linguistic patterns, emotional trajectories, and dramatic tension, we gain a nuanced understanding of Koltès' thematic preoccupations and narrative techniques. The variation in vocabulary richness across the plays highlights Koltès' adaptive linguistic style, tailored to the unique emotional demands of each narrative. The analysis of word frequency and emotional content illuminates his ability to weave complex psychological landscapes, using language not just as a vehicle for dialogue but as a tool for evoking atmosphere and tension. These insights affirm Koltès' mastery in exploring the human condition through a confluence of existential themes, stark realism, and poetic expression, reinforcing his significant contribution to modern theatre. By bridging the visceral and the intellectual, Koltès' work challenges audiences to confront the multifaceted realities of existence, positioning his plays as pivotal explorations of contemporary life and its discontents.

\subsection{Methodological Reflections}
Employing NLP for the analysis of literary texts, particularly dramatic works, presents a novel intersection of computational linguistics and literary scholarship. This methodology's strengths lie in its ability to process and analyze large volumes of text systematically, uncovering patterns and trends that might elude traditional close reading. For instance, sentiment analysis and emotion detection offer quantifiable metrics to gauge dramatic tension and emotional depth, providing objective grounding to subjective interpretations. While this approach offers valuable insights, it is important to acknowledge its limitations. The subtleties of literary language, including irony, metaphor, and other rhetorical devices, can present challenges for NLP tools and may lead to misinterpretations. Furthermore, the cultural and historical nuances embedded in literary texts require a nuanced understanding that purely computational methods might miss. To mitigate these issues, we supplemented the analysis with manual verification and interpretive analysis to ensure that the computational findings were contextualized within a broader literary framework. This combination of computational techniques and humanistic interpretation highlights the potential of interdisciplinary methods to enhance literary studies, while recognizing the inherent complexities of literary texts.

\bibliography{koltes_emotion}

\clearpage

\appendix

\begin{figure*}[!htb]
    \centering
    \includegraphics[width=\textwidth]{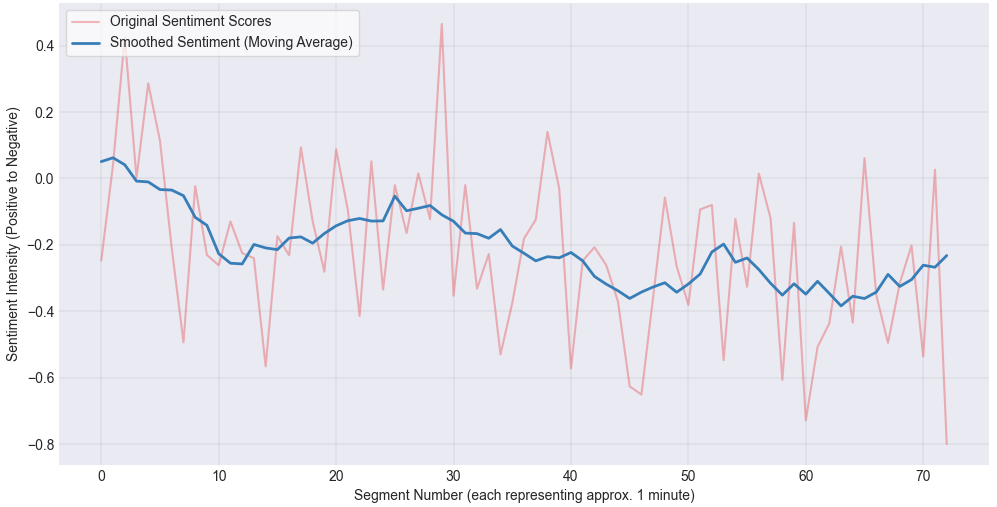}
    \caption{Sentiment Analysis Over Time in "Dans la Solitude des Champs de Coton"}
    \label{fig:sentiment-time-dans}
\end{figure*}

\begin{figure*}[!htb]
    \centering
    \includegraphics[width=\textwidth]{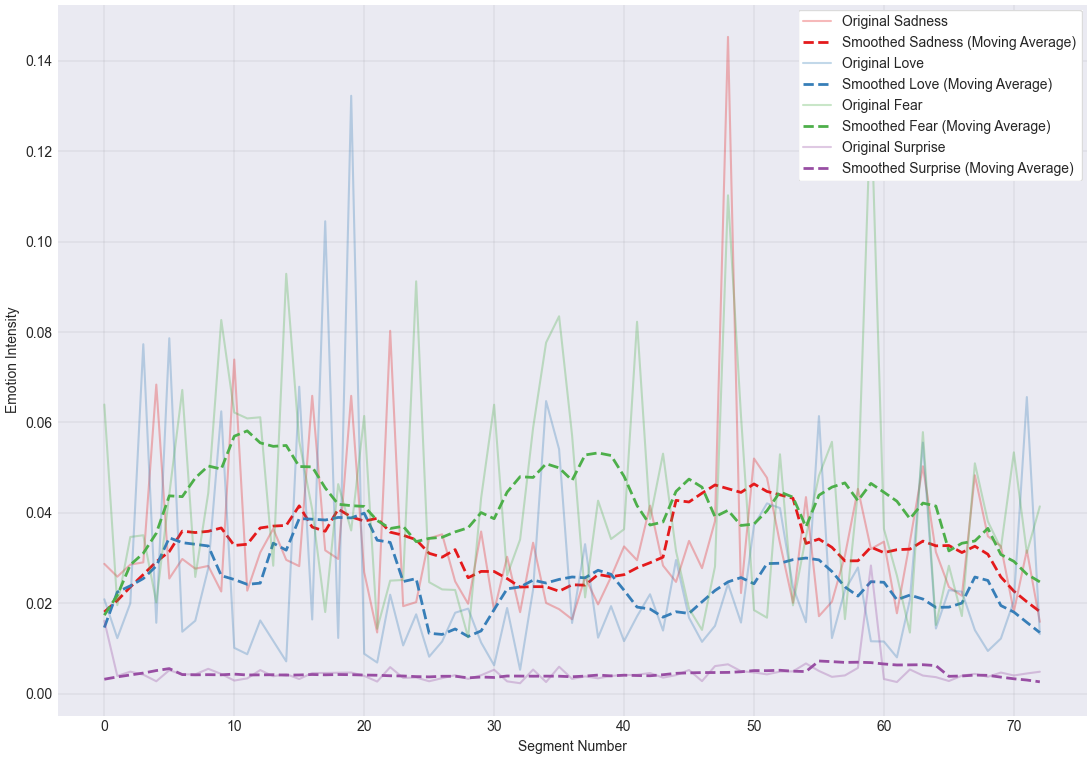}
    \caption{Emotion Analysis Over Time in "Dans la Solitude des Champs de Coton"}
    \label{fig:emotion-time-dans}
\end{figure*}

\begin{figure*}[!htb]
    \centering
    \includegraphics[width=0.8\textwidth]{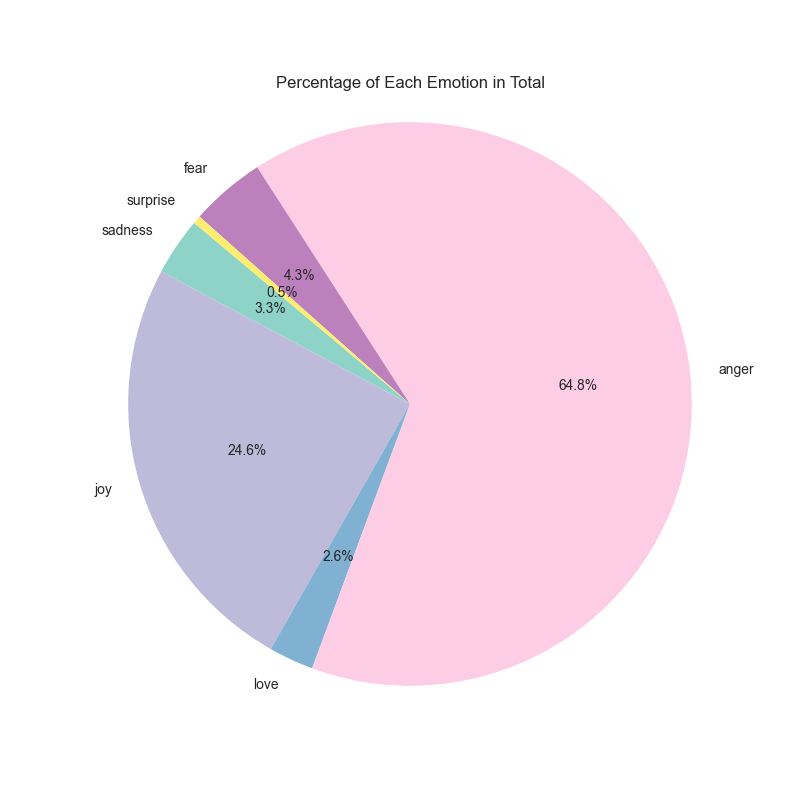}
    \caption{Percentage Distribution of Emotions in "Dans la Solitude des Champs de Coton"}
    \label{fig:emotion-percentage-dans}
\end{figure*}

\begin{figure*}[!htb]
    \centering
    \includegraphics[width=0.8\textwidth]{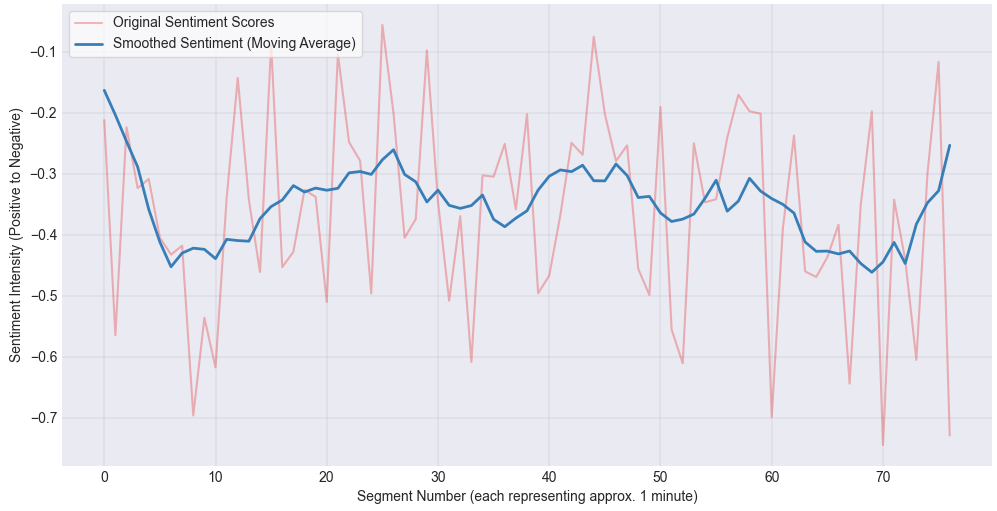}
    \caption{Sentiment Analysis Over Time in "La Nuit Juste Avant les Forets"}
    \label{fig:sentiment-time-nuit}
\end{figure*}

\begin{figure*}[!htb]
    \centering
    \includegraphics[width=0.8\textwidth]{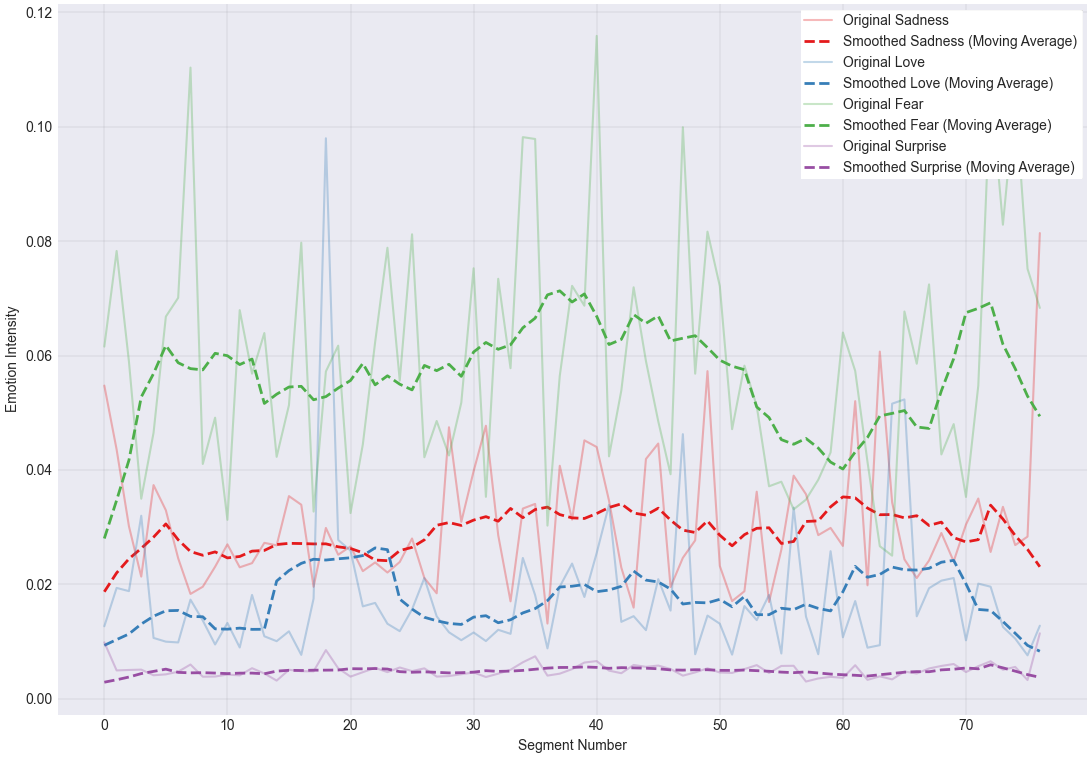}
    \caption{Emotion Analysis Over Time in "La Nuit Juste Avant les Forets"}
    \label{fig:emotion-time-nuit}
\end{figure*}

\begin{figure*}[!htb]
    \centering
    \includegraphics[width=0.8\textwidth]{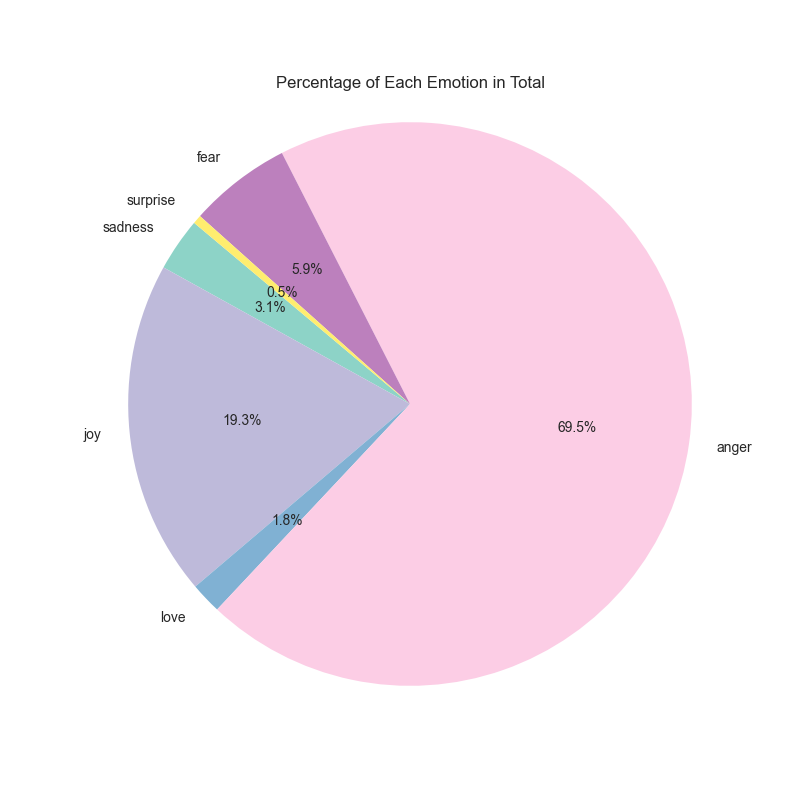}
    \caption{Percentage Distribution of Emotions in "La Nuit Juste Avant les Forets}
    \label{fig:emotion-percentage-nuit}
\end{figure*}

\begin{figure*}[!htb]
    \centering
    \includegraphics[width=0.8\textwidth]{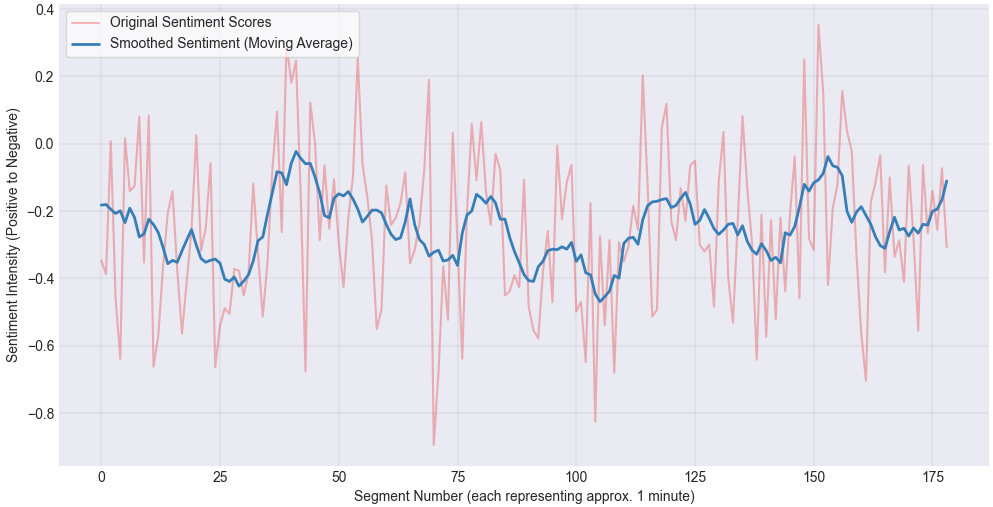}
    \caption{Sentiment Analysis Over Time in "Combat de Negre et de Chiens"}
    \label{fig:sentiment-time-combat}
\end{figure*}

\begin{figure*}[!htb]
    \centering
    \includegraphics[width=0.8\textwidth]{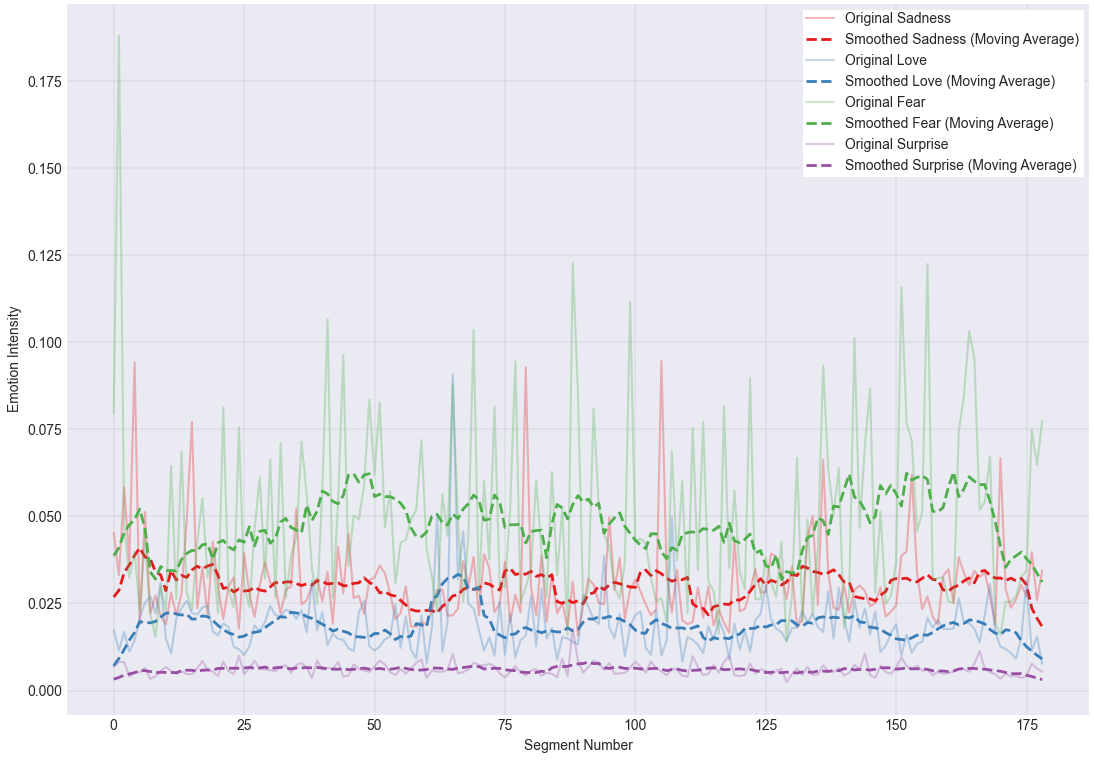}
    \caption{Emotion Analysis Over Time in "Combat de Negre et de Chiens"}
    \label{fig:emotion-time-combat}
\end{figure*}

\begin{figure*}[!htb]
    \centering
    \includegraphics[width=0.8\textwidth]{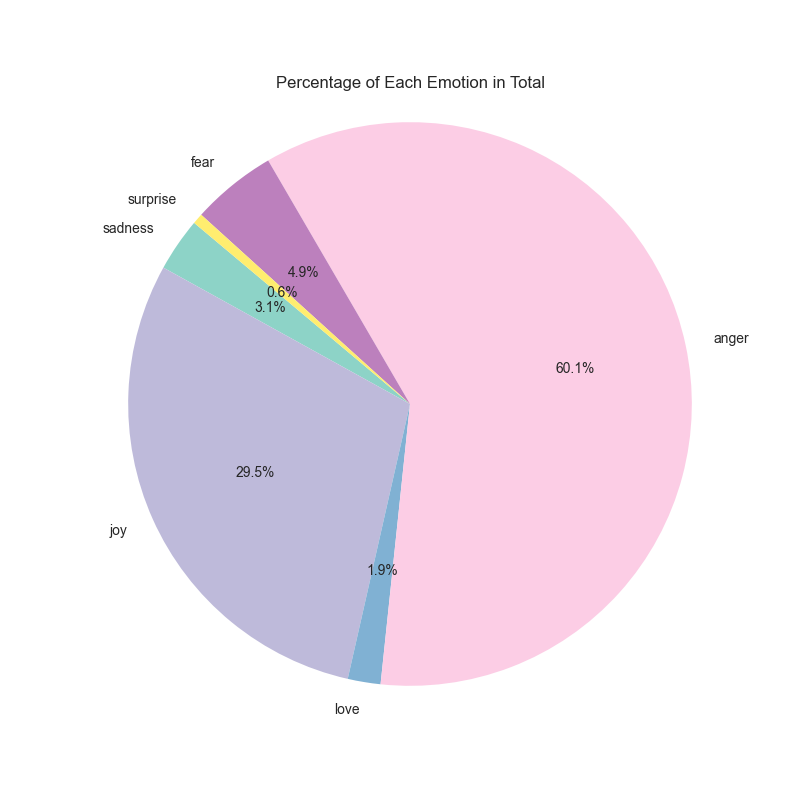}
    \caption{Percentage Distribution of Emotions in "Combat de Negre et de Chiens""}
    \label{fig:emotion-percentage-combat}
\end{figure*}

\end{document}